\documentclass[11pt]{article}
\PassOptionsToPackage{switch,mathlines}{lineno}
\usepackage{amsmath}       
\usepackage{algorithm}
\usepackage{amssymb} 
\usepackage{algpseudocode} 
\usepackage{times}
\usepackage{pdfpages}
\usepackage{latexsym}
\usepackage{booktabs} 
\usepackage{multirow}
\usepackage{xspace}
\usepackage{float}
\usepackage{CJKutf8}
\usepackage{ulem}
\usepackage{mdframed}
\usepackage{xcolor}
\usepackage{fvextra} 
\usepackage{mdframed}
\usepackage{cuted}  
\usepackage{array}  

\fvset{
  breaklines=true,
  breakanywhere=true,
  breaksymbolleft={},
  breaksymbolright={},
  breaksymbolsep=0pt
}

\newmdenv[
  linewidth=0.8pt,
  linecolor=blue!60!black,
  backgroundcolor=white,
  roundcorner=2pt,
  innertopmargin=8pt, innerbottommargin=8pt,
  innerleftmargin=8pt, innerrightmargin=8pt,
  frametitlerule=true,
  frametitlebackgroundcolor=blue!8
]{promptbox}

\newenvironment{PromptBox}[1]{%
  \begin{promptbox}[frametitle={#1}]%
}{%
  \end{promptbox}%
}

\usepackage[final]{acl}

\usepackage{times}
\usepackage{latexsym}

\usepackage[T1]{fontenc}

\usepackage[utf8]{inputenc}

\usepackage{microtype}

\usepackage{inconsolata}

\usepackage{graphicx}

\title{Towards semantic reliable clinical QA: Query pipeline optimization for cancer patient question answering systems}

\author{MaoLin He$^\heartsuit$, Rena Gao$^{\heartsuit, \clubsuit}$, Mike Conway$^\heartsuit$, Brian E. Chapman$^\spadesuit$  \\
  $^\heartsuit$School of Computing and Information Systems, University of Melbourne \\
   $^\spadesuit$Health Data Science and Biostatistics, University of Texas Southwestern Medical Center \\
   $^\clubsuit$ Keeta AI \\
  \texttt{mlhinaus@gmail.com}, \texttt{wegao@student.unimelb.edu.au} \\
  \texttt{mike.conway@unimelb.edu.au},
  \texttt{Brian.Chapman@utsouthwestern.edu}
}

\begin{document}
\maketitle
\begin{abstract}
Large Language Models (LLMs) show promise in medical Question-Answering (QA) but suffer from hallucinations that jeopardize patient safety. While Retrieval-Augmented Generation (RAG) mitigates this by grounding outputs in external evidence, existing pipelines struggle with the complex, rapidly evolving nature of oncology. We present \textbf{CoMeta}, a three-level controllable metadata-aware framework optimized for Cancer Patient QA (CPQA). We introduce Clinical Hybrid Semantic-Symbolic Document Retrieval (CHSDR), which synergizes real-time Boolean search via NCBI E-Utilities with semantic retrieval to overcome metadata blindness. Additionally, we propose \textbf{S}emantic \textbf{E}nhanced \textbf{O}verlap \textbf{S}egmentation (SEOS) to prevent contextual fragmentation. Our results demonstrate that CHSDR significantly improves retrieval performance, CoMeta improved the answer accuracy of Claude-3-haiku by 5.24\% over chain-of-thought prompting and about 3\% over a naive RAG
setup.This study highlights the importance of domain-specific query optimization in realizing the full potential of RAG and provides a robust framework for building more
reliable CPQA systems.
\end{abstract}

\section{Introduction}
Large Language Models (LLMs) have shifted the paradigm of online information seeking from traditional search engines to conversational agents. While promising for Question-Answering (QA), their reliance on internal parameters renders them prone to hallucination—generating fluent but factually incorrect outputs \cite{ji2023survey}. This issue is acute in the \textbf{clinical domain}, where accuracy directly impacts patient safety \cite{umapathi2023med}. To mitigate this, Retrieval-Augmented Generation (RAG) \cite{lewis2020retrieval} integrates external evidence to ground LLM responses. Especially in medical QA tasks where queries are knowledge intensive, LLMs excel as generators rather than knowledge databases \cite{truhn2023large}. 

To enhance medical QA, prior studies \cite{jeong2024improving, liang2025rgar} broadly use RAG with domain-specific adaptation of retrieval strategies. However, they overlook the query pipeline, the foundational mechanism for query-to-evidence mapping that underpins all retrieval strategies. In Cancer Patient QA (CPQA), these pipelines cause systematic retrieval failures.
First, retrieval reliability face the \textbf{staleness-semantic dilemma}: the rapid evolution of oncology \cite{landhuis2016scientific} means that standard query pipelines (e.g., Dense or BM25), built upon static and metadata-blind indexes, risk surfacing outdated and thus clinically unreliable evidence, while real-time, metadata-aware interfaces like E-Utilities \cite{kans2024entrez} are semantically brittle to informal patient queries.
Second, a \textbf{retrieval-depth paradox} stems from the asymmetric utility of publication types. We argue that reviews require full-text retrieval to capture high-level therapeutic syntheses \cite{wang2025trustworthy}, while primary research articles often benefit from abstract-only retrieval to avoid methodological noise. Most pipelines apply uniform retrieval depth across article types, either starving the model of synthesis or overwhelming it with clutter.
Third, semantic representation is undermined by \textbf{contextual fragmentation}: prior encoder-agnostic segmentation (pre-defined length \cite{liu2022llamaindex} or lexical \cite{hearst1997text}) severs clinical qualifiers (e.g., specific mutation criteria) from therapeutic statements, yielding recommendations that appear evidence-based yet lack essential medical constraints.

We address these failures with \textbf{CoMeta}, a specialized RAG framework for CPQA that enforces \emph{controllability} across three pillars:
1) Robustness against the staleness-semantic dilemma, ensuring stable performance across expert queries and informal patient narratives; 2) Evidence Metadata-awareness, enabling time filtering and adaptive retrieval depth based on publication types; 3) Relational Integrity, utilizing encoder-aware segmentation to safeguard clinical logic.
To support this architecture, we curate the Cancer-related Medical QA (CMMQA) dataset and its "Clinical Narrative Variant" to simulate real-world patient interactions \footnote{All codes and datsets can be found via: \url{anonymous.4open.science/r/COMP90005-E51E/README.md}.}. 
Our specific contributions are:
\begin{itemize}
    \item We propose the Clinical Hybrid Semantic-Symbolic Document Retrieval (CHSDR) method. To our knowledge, this is the first pipeline in the clinical RAG domain to combine real-time Boolean search (via E-Utilities with LLM-rewritten queries) with MedCPT \cite{jin2023medcpt} semantic search.
    \item We introduce \textbf{S}emantic \textbf{E}nhanced \textbf{O}verlap \textbf{S}egmentation (SEOS), a novel text splitter that integrates sentence semantics and embedding model characteristics while utilizing chunk overlap to preserve critical context.
    \item We provide the first comparative analysis of NCBI sources for CPQA, demonstrating that PMC review articles possess significantly higher retrieval value than non-review PMC papers among clinical cancer QA datasets.
\end{itemize}

\section{Related Work}
\subsection{Retrieval in Medical RAG}

Medical RAG systems rely on their query pipelines, which map user queries to evidence. The predominant paradigm is dense retrieval, leveraging domain-specific embedding models for vector-similarity search in offline index \cite{miao2025improving}. Recent advances have shifted from general-purpose embeddings to domain-specific retrievers like MedCPT \cite{jin2023medcpt}.
Advanced retrieval strategies, including hybrid search that integrates sparse methods like BM25 \cite{robertson2009probabilistic} for lexical-semantic complementarity \cite{xiong-etal-2024-benchmarking,xu2026selfcorrectingrag}, adaptive retrieval that dynamically selects retrieval timing \cite{jeong2024improving} and recursive search that iteratively refine queries through feedback loops \cite{liang2025rgar}, are essentially optimizations built upon static-index pipelines. 

The reliance on offline indexing restricts metadata filtering (e.g., by publication date or study type), incurs substantial preprocessing and storage overhead (e.g., $>$400GB for PubMed indices \cite{jeong2024improving}), and introduces timeliness gaps where newly published studies may remain inaccessible in critical decision windows. These issues are acute in oncology, where clinical standards evolve rapidly.  NCBI’s E-Utilities offers a potential remedy through real-time access and metadata filtering, but its utility is hindered by a term-centric matching paradigm. Unlike BM25, E-Utilities lacks mechanisms like length normalization or term weighting, making retrieval highly sensitive to query formulation and ill-suited for natural language inputs without significant adaptation.

Thus, existing systems are implicitly forced to trade off semantic robustness against retrieval controllability. 
We instead explore a different design paradigm. Rather than further optimizing static-index pipelines, CHSDR mitigates these limitations by integrating E-utilities as a live, metadata-aware sparse backend. This design is orthogonal and complementary to prior RAG optimizations (e.g., MedRAG, Self-BioRAG, Adaptive RAG), enabling seamless integration with prior works.

\subsection{Semantic Representation in RAG}
\label{existing text splitters}
Dense retrieval efficacy hinges on precise semantic representation, which is governed by two coupled factors: document segmentation and text encoding \cite{gao2023retrieval}. While recent medical RAG systems extensively optimize retriever configuration \cite{xiong-etal-2024-benchmarking, tang2024multihop}, document segmentation, the foundational step determining input granularity, remains critically under-explored.

Current approaches largely rely on heuristic strategies that compromise clinical context. Naive fixed-length chunking \cite{jeong2024improving} frequently truncates sentences mid-thought, severing syntactic dependencies. Even sentence-aware methods like LlamaIndex’s SentenceSplitter \cite{liu2022llamaindex}, while preserving sentence integrity, impose rigid window sizes that fail to adapt to the variable information density of medical narratives. Similarly, TextTiling \cite{hearst1997text} infers topic boundaries via lexical overlap, but its surface-level matching struggles with the high synonymy and complex semantic shifts inherent in biomedical literature.

The fundamental limitation of these approaches is twofold. First, they function as ``semantic-blind'' preprocessing steps, leading contextual fragmentation. Second, they neglect the interaction between chunk size and encoder performance, ignoring evidence that optimal segmentation is highly dependent on the specific embedding model used \cite{gao2023retrieval}.

\section{Methods}
\subsection{Dataset and Evaluation}
\begin{figure}[htbp] 
    \centering
    \includegraphics[width=0.5\textwidth]{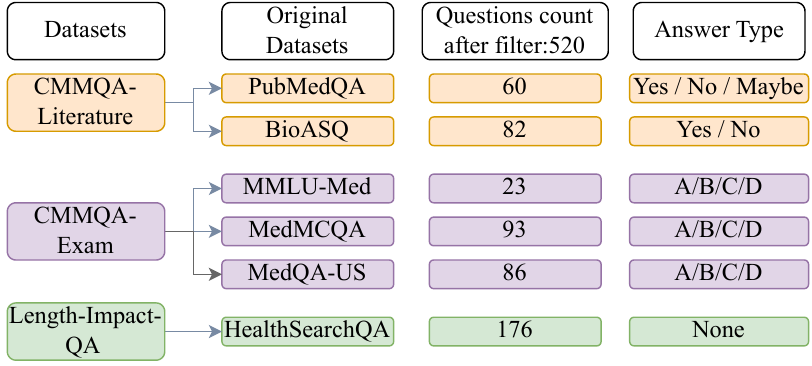}
    \caption{Description of filtered cancer QA datasets used in this study. }
    \label{fig:ds}
\end{figure}

We applied a MeSH-based filter to HealthSearchQA \cite{singhal2023towards}
and the MIRAGE benchmark \cite{xiong-etal-2024-benchmarking} that includes PubMedQA \cite{jin2019pubmedqa}, BioASQ \cite{tsatsaronis2015overview}, MedQA \cite{jin2021disease}, MedMCQA \cite{pal2022medmcqa}, the medical subsets in MMLU \cite{hendrycks2020measuring}. 
Specifically, we use all terms and synonyms under the 'neoplasm' MeSH subtree to identify questions strictly pertaining to cancer as a disease entity. While not exhaustive, it ensures a focus on core disease-related queries. We refer to this filtered collection as the Cancer-related Medical QA Dataset (CMMQA)  (Figure \ref{fig:ds}). Then, we use LLama-3-70b to rewrite the CMMQA queries into non-expert clinical narratives (mainly simulating patients), creating a derived dataset we term the ``Clinical Narrative Variant". This variant serves as a challenging testbed to evaluate both retrieval stability and QA accuracy across distinct query modalities (standard vs. narrative). 

Our \textit{document} retrieval evaluation focuses on PubMedQA and BioASQ because these two datasets provide gold-standard citations (PMIDs) but others lack source annotations. 
For \textit{passage} retrieval evaluation, we utilized synthetic Query-Answer pairs generated from PubMed Abstracts, PMC Full Texts and medical textbooks \cite{jin2021disease} as a validation set. Retrieval Performance is measured at a cutoff of K=10 using three standard metrics: Hit Rate (Hit@K), representing the percentage of queries with at least one relevant document retrieved; Mean Reciprocal Rank (MRR@K), assessing the ranking quality of the first relevant document; and Recall (Recall@K), quantifying the coverage of gold-standard evidence. Besides, we report Hit0,  the count of queries yielding zero results. This metric serves as a critical indicator of retrieval robustness, highlighting system failures in handling complex or restrictive queries.






\subsection{Controllable Query Pipeline Design}  
\begin{figure}[htbp]
    \includegraphics[width=1\linewidth]{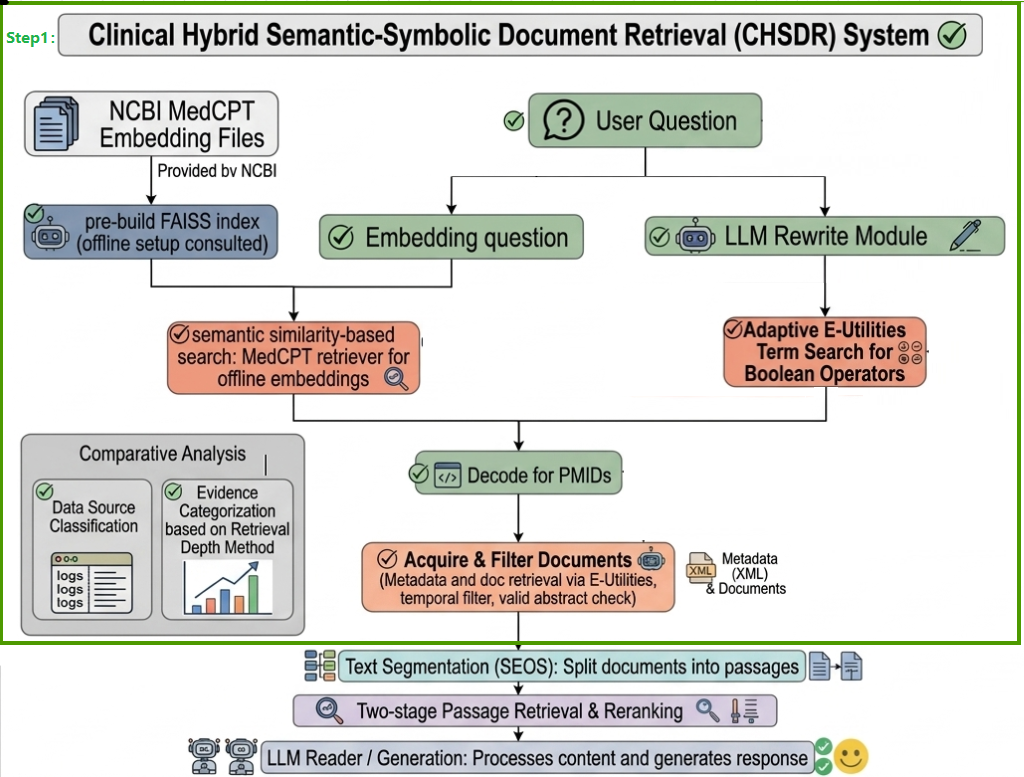}
    \caption{The CHSDR employs dual retrieval strategies, then downloads and filters candidate documents. After document Retrieval, next steps and comparative analyses are conducted}
    \label{fig:CQP}
\end{figure}

 Our target is real clinical settings that do not maintain a local PubMed mirror but do require metadata‑aware, up‑to‑date evidence access. To ensure clinical rigor across the retrieval lifecycle, we design a hierarchical query pipeline (Figure \ref{fig:CQP}) that prioritizes \emph{controllability} at both the document and passage levels.

\subsubsection{Clinical Hybrid Semantic-Symbolic Document Retrieval (CHSDR)}

While BM25 paired with dense retrieval offers strong lexical complementarity, it relies on static indices that are disconnected from real-time updates. In contrast, 
CHSDR (stage-1 in figure \ref{fig:CQP}) enables metadata control and robust retrieval across various queries (standard vs. clinical narrative), overcoming \textbf{staleness-semantic dilemma} and \textbf{retrieval-depth paradox}. The system design comparison table are detailed in Appendix \ref{sec:bm25_vs_eutils}. 

\paragraph{Boolean-constrained retrieval with adaptive query execution (Adapt-E)} Raw consumer questions contain orthographic and grammatical errors and vary in length \cite{lu2019spell, abacha2019bridging}, which are incompatible with E-Utilities. 
CHSDR thus use a LLM-based rewriter that performs not only query error correction and text normalization \cite{lu2019spell}, but also patient intent analysis, clinical abstraction (i.e.,mapping narrative descriptions into standardized medical concepts and PICO elements), boolean expression and temporal constraint generation. To ensure robustness, we design a \textit{adaptive query execution}: generated outputs are executed progressively from strict boolean conjunctions to clinical abstraction of raw queries to relaxed boolean formulations, until sufficient documents are retrieved.

\paragraph{Hybrid semantic–symbolic retrieval with metadata-aware control}
We integrate symbolic search with semantic retrieval via Reciprocal Rank Fusion (RRF) \cite{cormack2009reciprocal}, further overcoming the term-centric limitations of E-Utilities. We adopt MedCPT \cite{jin2023medcpt} for semantic document retrieval, because it is trained for query-article retrieval and achieves a balance between performance and efficiency. Both retrieval streams return PubMed identifiers (PMIDs), which serve as a unified document key.
All candidate documents and their associated metadata are subsequently accessed via E-Utilities. 

\paragraph{Metadata Utilization}
Retrieved documents are augmented with metadata parsed from 
E-Utilities XML responses. 
We exploit three main fields: 
(1) \textit{Publication Type} (\texttt{pt}), which classifies documents into 
PubMed abstracts (D1), PMC reviews full-text (D2), or non-review PMC papers (D3), enables empirical analysis of retrieval depth across publication types; 
(2) \textit{Publication Date} (\texttt{dp}), which can be used to enforces recency filters and prioritizes more recent evidence when conflicts arise; 
(3) \textit{Abstract Availability}, which excludes records lacking valid 
abstracts to ensure data quality.

.

\subsubsection{Passage Retrieval and Text Splitter} 

\paragraph{Semantic Enhanced Overlap Segmentation (SEOS)}
To address the twofold limitation in section \ref{existing text splitters}, we propose SEOS (pseudo-code in Appendix \ref{pseudo}). SEOS is inspired by TextTiling \cite{hearst1997text}, which identifies segmentation boundaries via local cohesion shifts, but extends it through following key innovations. First, We replace bag-of-words with domain-specific dense embeddings in text representation, enabling robust handling of medical terminology and discourse relations in clinical studies. Second, instead of brittle similarity thresholds, SEOS determines the optimal partition cardinality (N) from a target token budget, selecting the top-N semantic minima as breakpoints.The budget is adaptively configured to align with the effective input granularity observed for different embedding models, as prior work has shown retrieval performance to be sensitive to chunk length \cite{gao2023retrieval,wang2019multi}. 
Third, to mitigate contextual fragmentation, SEOS incorporates adaptive sentence overlap determined based on semantic continuity around each breakpoint, preserving unresolved semantic dependencies. Furthermore, neighboring chunk identifiers are explicitly stored, enabling local context recovery when evidence spans multiple segments.

\paragraph{Dense retrieval and reranking for passage matching}
The training objective of MedCPT is optimized for matching queries to PubMed abstract, but our passage retrieval corpus mainly is PMC full text that exhibit different discourse structures and information densities. This distributional shift from abstracts to full-text passages necessitates encoders with broader generalization capabilities. Therefore, we select top-performing embedding models from the MTEB benchmark \cite{muennighoff2022mteb}. For the subsequent reranking stage, we employ cross-encoder architectures to capture the fine-grained, non-linear semantic interactions that bi-encoders often miss \cite{jiang2023boot}. To evaluate the trade-off between domain specialization and general semantic reasoning, we only select representative ones: 1) bge-reranker-v2-m3 \cite{li2023making}, a general-purpose model with superior performance; and 2) MedCPT-reranker \cite{jin2023medcpt}, a domain-specific reranker pre-trained on large-scale biomedical literature.

\section{Results}
We compare our work with two settings: 1) LLM with COT, a retrieval-free approach relying 
solely on parametric knowledge. Chain-of-Thought (CoT) \cite{wei2022chain} encourages the LLM to perform step-by-step reasoning to improve the quality of the generated answers. 2) LLM with naive RAG, using the whole text of the Top 5 relevant documents retrieved by the MedCPT to directly guide LLM to generate an answer. 

\subsection{Analysis in Document retrieval} 

\subsubsection{Component Ablations of CHSDR} 
In Table \ref{tab:main_results}, we conduct the ablation study of CHSDR on retrieval performance. Results indicate:  
\noindent\emph{1. Impacts of Clinical Narrative.} 
While shifting to \textit{Clinical Narrative} inputs universally degrades performance, the hybrid model maintains robustness. This suggests that when the input text is noisy, the semantic understanding provided by the Hybrid system becomes the critical mechanism for retrieval resilience.
\noindent\emph{2. the Hybrid Dilemma.}
The \textbf{Hybrid} model achieves best performance on BioASQ, outperforming \textbf{Adapt-E} method. This indicates that semantic retrieval (MedCPT) successfully recalls relevant documents that are not retrieved by sparse retriever. However, on PubMedQA, the Hybrid approach slightly underperforms the symbolic-only Adapt-E method in the \textit{Standard Narrative}. We attribute this anomaly to query specificity:
PubMedQA questions are often derived directly from abstract sentences, creating a bias toward exact lexical matching. 
\noindent\emph{3. the "Zero-Hit" Barrier.} Across most datasets, the limitations of sparse search are evident in the \textbf{E-utils} baseline. In contrast, \textbf{Adapt-E} method (LLM Rewriter + Adaptive Fallback) improve retrieval performance across all datasets, particularly in the \textit{Clinical Narrative} setting where it help the system to recover from the baseline's total failure (high Hit0). This demonstrates the accuracy and robustness of rewriting module, which acts as a crucial bridge between patient queries and biomedical literature. As demonstrated by the representative failure-recovery illustration in Appendix \ref{sec:case_study}, the LLM rewriting module does not rely on a single ``perfect'' rewrite. Instead, it generates a hierarchy of Boolean candidates (Strict $\rightarrow$ Relaxed) and executes them sequentially via our Adaptive Fallback mechanism, enabling collect sufficient evidence even for complex queries.

\begin{table*}[t]
    \centering
    \small 
    \setlength{\tabcolsep}{3.5pt} 
    \begin{tabular}{ll cccc c cccc}
        \toprule
        & & \multicolumn{4}{c}{\textbf{Standard Narrative}} && \multicolumn{4}{c}{\textbf{Clinical Narrative}} \\
        \cmidrule(lr){3-6} \cmidrule(lr){8-11}
        \textbf{Dataset} & \textbf{Method} & Hit@10 & MRR@10 & Recall@10 & Hit0$\downarrow$ && Hit@10 & MRR@10 & Recall@10 & Hit0$\downarrow$ \\
        \midrule
        \multirow{4}{*}{\textbf{PubMedQA}} 
          & E-utils       & 41.67 & 38.06 & 41.67 & 22 && 0.00 & 0.00 & 0.00 & 55 \\
          & \textbf{Adapt-E}& \textbf{48.33} & \textbf{44.72} & \textbf{48.33} & \textbf{0}  && 8.33 & 6.04 & 8.33 & \textbf{1} \\
          & MedCPT        & 10.00 & 9.17  & 10.00 & 0  && 3.33 & 2.50 & 3.33 & 0 \\
          & Hybrid        & 46.67 & 33.25 & 46.67 & 0  && \textbf{10.00} & \textbf{5.28} & \textbf{10.00} & 0 \\
        \midrule 
        \multirow{4}{*}{\textbf{BioASQ}} 
          & E-utils       & 52.44 & 35.49 & 28.87 & 18 && 1.22 & 1.22 & 1.22 & 76 \\
          & \textbf{Adapt-E}& 65.85 & 41.94 & 41.61 & \textbf{0}  && 50.00 & 28.93 & 33.32 & \textbf{1} \\
          & MedCPT        & 63.41 & 42.39 & 36.50 & 0  && 41.46 & 21.43 & 21.70 & 0 \\
          & Hybrid        & \textbf{80.49} & \textbf{49.74} & \textbf{51.21} & 0  && \textbf{60.98} & \textbf{35.36} & \textbf{40.07} & 0 \\
        \bottomrule
    \end{tabular}
    \caption{Ablation study on retrieval performance. PubMedQA and BioASQ have introduced in Figure \ref{fig:ds}. \textbf{Adapt-E} refers to our Boolean-constrained retrieval with adaptive query execution. \textbf{E-utils} refer to E-utilizes.}
    \label{tab:main_results}
\end{table*}

\subsubsection{Study on Evidence Sources and Depth}
\begin{figure}[htbp]
    \centering
    \includegraphics[width=0.5\textwidth]{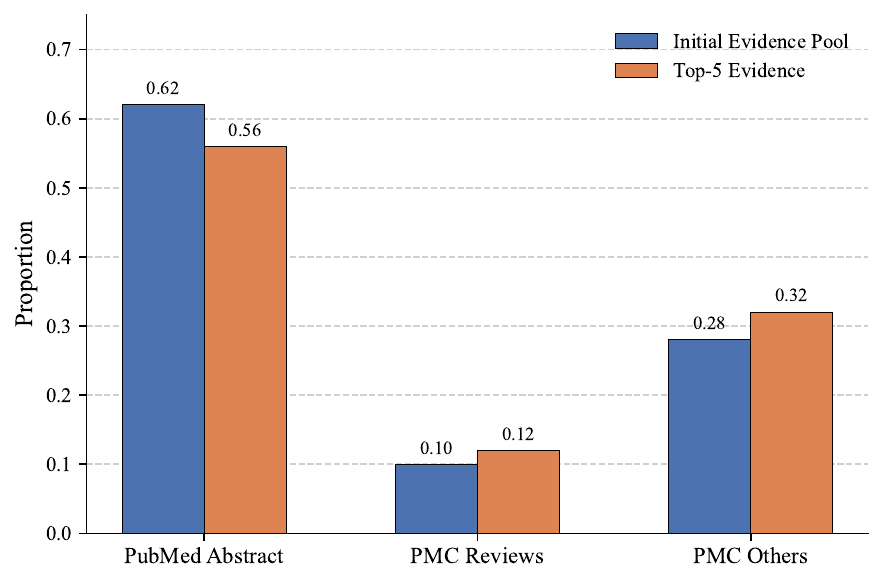}
    \caption{Distribution comparison between Initial Evidence Pool and Top-5 Evidence when CHSDR's Retrieval Source involving PubMed Abstract, PMC Reviews and PMC Others}
    \label{fig:5}
\end{figure}

To thoroughly investigate the comparative value of different NCBI literature sources for CPQA, we evaluate both their retrieval distribution (Figure \ref{fig:5}) and their downstream impact on generation performance (Table \ref{tab:ds-n-cancer-question-results}). We sampled 100 questions from CMMQA via stratified sampling for this analysis.

Figure \ref{fig:5} reveals two key insights regarding the comparative value of NCBI sources. First, while PubMed abstracts dominate both the initial retrieval pool and the top-5 evidence due to broader coverage (23.9M indexed citations vs. 8M PMC full-texts), their share decreases in the final top-5 evidence. This suggests that full-text sources provide richer, more discriminative content for CPQA, highlighting the value of integrating complementary data sources. Second, among the full-text sources, the relative growth of PMC Reviews outpaces that of other PMC papers (from 0.1/0.28 to 0.12/0.32), indicating that review articles possess a higher retrieval value for question answering.
\begin{table}[H]
    \centering
    \small 
    \setlength{\tabcolsep}{6pt} 
    \resizebox{\columnwidth}{!}{
    \begin{tabular}{l cccc}
        \toprule
        \textbf{Source} & \textbf{Acc} & \textbf{Prec} & \textbf{Rec} & \textbf{F1} \\
        \midrule
        D1 (Abstracts)       & 44.00 & 41.12 & 40.12 & 39.14 \\
        D1 + D2 (Reviews)    & \textbf{46.00} & \textbf{43.78} & \textbf{38.13} & \textbf{38.77} \\
        D1 + D2 + D3 (Other) & 46.00 & 40.12 & 33.86 & 35.33 \\
        \bottomrule
    \end{tabular}
    }
    \caption{Impact of data sources on retrieval performance (Negative Cancer QA Dataset). D1: PubMed Abstracts; D2: PMC Reviews; D3: Other PMC articles.}
    \label{tab:ds-n-cancer-question-results}
\end{table} 

Table~\ref{tab:ds-n-cancer-question-results} further shows these patterns. Integrating PMC reviews with PubMed abstracts (D1 + D2) improves accuracy from 44.00\% to 46.00\%, while further including non-review full-texts (D1+D2+D3) maintains accuracy but degrades Precision, Recall, and F1.
This divergence is mechanistically coherent: review articles synthesize 
findings across studies, matching the broad scope of patient queries. In contrast, non-review full-texts are typically context-specific; their core clinical outcomes can be represented by their abstracts, and retrieving their full texts introduces noise that overwhelms the model.

These empirical findings validate CoMeta's solution to the \textbf{retrieval-depth paradox}: before passage retrieval, the metadata-aware pipeline should calibrates retrieval depth based on publication type. This step is a core component of CHSDR and distinguishes CoMeta from pipelines that treat all retrieved documents uniformly.

\subsection{Analysis in Passage Retrieval}
\subsubsection{Comparison of Text chunking Strategies}
Table~\ref{tab:ts-n-cancer-question-results} evaluates the robustness of SEOS across passage retriever, including lexical retrieval (BM25), domain-specific dense retrieval (MedCPT), and sentence-level embedding models (PubMedBERT-matryoshka). Results demonstrates the SEOS method's superiority in text segmentation, outperforming fixed-parameters strategies. These fixed-length baselines represent the standard chunking strategies widely adopted across current mainstream medical RAG frameworks, including MedRAG, Self-BioRAG, and LlamaIndex. Comparisons with new advanced semantic chunking strategies will be included in future work.

\begin{table}[H]
    \centering
    \small 
    \setlength{\tabcolsep}{6pt} 
    \resizebox{\columnwidth}{!}{
    \begin{tabular}{l ccc}
        \toprule
        & \multicolumn{3}{c}{\textbf{Retriever Accuracy (\%)}} \\
        \cmidrule(lr){2-4}
        \textbf{Splitter Configuration} & \textbf{PubMedBERT} & \textbf{BM25} & \textbf{MedCPT} \\
        \midrule
        512 (Overlap 0)    & 46 & 20 & 22 \\
        512 (Overlap 32)   & 52 & 18 & 24 \\
        512 (Overlap 128)  & 42 & 16 & 22 \\
        \textbf{SEOS (Ours)} & \textbf{54} & \textbf{36} & \textbf{38} \\
        \bottomrule
    \end{tabular}
    }
    \caption{Retrieval accuracy across different text segmentation strategies on the Passage Retrieval Evaluation Dataset. SEOS outperforms fixed-size chunking.}
    \label{tab:ts-n-cancer-question-results}
\end{table}

The SEOS method excels by preserving natural and meaningful text boundaries based on semantic similarity and its variations, which enhances the retrievers' ability to locate relevant information. Its advantages also include sentence overlap and automatic chunk-size adjustment tailored to the embedding model. Both Pubmedbert-matryoshka and MedCPT benefited from automatic chunk-size adjustment. The corresponding chunk-size adjustment rule obtained from research indicates that 128-word chunks with 32-word overlaps optimize BERT-based models for QA tasks \cite{wang2019multi}. This finding can also be shown by the 512Overlap32's top performance among fixed-length strategies. Despite a 512 chunk size, the actual retrieval text space is roughly 128, with metadata integrated consuming the remainder of the chunk, typically around 384 tokens.

\subsubsection{Selection of Retrievers and Rerankers}

\begin{figure}[htbp]
    \centering
    \includegraphics[width=0.5\textwidth]{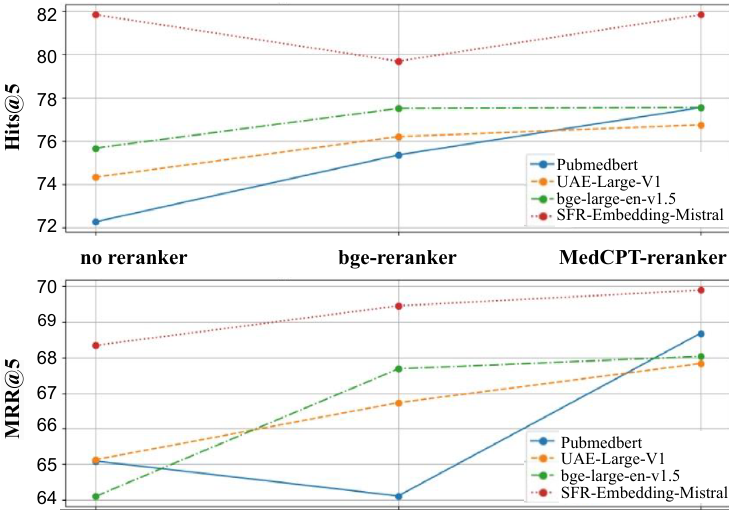}
    \caption{Performance of different reranker-retriever (reranker-embedding) pairs measured by Hit@5 (top) and MRR@5 (bottom).}
    \label{fig:4}
\end{figure}

While MedCPT is suited for document retrieval due to its training on query-article pairs, it does not utilize a sentence-transformer structure, which may limit its precision for shorter texts. Therefore, to find suitable embedding-reranker pairs for finer-grained passage retrieval, we designed a systematic retrieval evaluation based on LlamaIndex \cite{liu2022llamaindex}, a framework for building RAG systems. Specially, we used E-Utilities to query "cancer" to retrieve PubMed abstracts and PMC full-text articles as the text corpus for building an evaluation dataset, then used LLMs to generate pairs (query, context) from each chunk of the prepared text corpus, ensuring this evaluation was suitable for all data sources. In the experiment, we evaluated retrieval performance using Hits@5 and MRR@5, which aligns with the practical constraints of RAG systems, where the limited context window of the LLM generator requires a focus on retrieving the most relevant chunks \cite{tang2024multihop}. Results are shown in Figure \ref{fig:4}.

\noindent\textbf{Domain Specialization Matters:} Pubmedbert-matryoshka , despite its smaller size and absence from the MTEB leaderboard, achieved the second-best performance when paired with the MedCPT-reranker. This suggests that the size of the embedding model is not the only determinant of effectiveness and that domain-specific fine-tuning or training can significantly improve performance by leveraging domain-specific understanding \cite{xiong-etal-2024-benchmarking}. The importance of domain-specific features is also demonstrated by the MedCPT-reranker outperforming the general domain reranker in enhancing retrieval across all embedding models. 

\noindent\textbf{Model Compatibility Matters:}
PubMedBERT-Matryoshka performed poorly without a reranker but benefited substantially from the MedCPT Reranker. This synergy likely stems from the MedCPT Reranker being trained on negative samples from the MedCPT retriever (derived from Pubmedbert), aligning the reranker more effectively with PubMedBERT’s embedding space and enabling it to capture complementary information. Meanwhile, the BGE Reranker enhances the performance of BGE Embedding in terms of hits, also suggesting the importance of compatibility and complementarity between the embedding model and the reranker. However, the observed performance decline when BGE Reranker is paired with incompatible embeddings highlights the risks of mismatched reranker-embedding combinations. 
If a reranker cannot align with the embedding space or provide complementary semantic insights, it can fail to capture semantic nuances or introduce noise, leading to performance degradation. 
In conclusion, while rerankers can enhance retrieval, selecting compatible reranker-embedding combinations is crucial.

\subsection{Discussion}
\noindent\textbf{Overall Performance.}
As shown in Table \ref{tab:final}, CoMeta outperforms Naive RAG by 2.91\% on average and LLM+CoT by 5.24\% on CMMQA. These gains demonstrates the effectiveness of our query pipeline optimization, which includes two complementary retrieval improvements: 
(i) CHSDR tackle \textbf{staleness-semantic dilemma} and \textbf{retrieval-depth paradox} through LLM-based query rewriting, adapative query execution, and metadata-aware hybrid search. (ii) SEOS and two-stage retrieval reduces semantic noise, ensuring that the model attends to most relevant, semantically segmented passages.

\begin{table}[H]
    \centering
    \small 
    \setlength{\tabcolsep}{3.5pt} 
    \resizebox{\columnwidth}{!}{
    \begin{tabular}{l ccccc c}
        \toprule
        \textbf{Method} & \textbf{MMLU} & \textbf{MedQA} & \textbf{MedMC} & \textbf{PMQA} & \textbf{BioASQ} & \textbf{Avg} \\ 
        \midrule
        LLM + CoT       & 78.26 & 68.60 & 65.59 & 45.00 & 80.49 & 67.15 \\ 
        Naive RAG       & \textbf{82.61} & 67.44 & 65.59 & 56.67 & 81.71 & 69.48 \\ 
        \textbf{CoMeta}   & \textbf{82.61} & \textbf{69.77} & \textbf{68.82} & \textbf{65.00} & \textbf{81.71} & \textbf{72.39} \\ 
        \bottomrule
    \end{tabular}
    }
    \caption{Performance comparison on CMMQA datasets. }
    \label{tab:final}
\end{table}

\noindent\textbf{Why Average Accuracy Understates CoMeta's Contribution.}
The average gain of 2.91\% over Naive RAG understates our contribution due to three limitations: 
\noindent\textit{(1) Blindness to ceiling effect of retrieval impact}
The identical performance on MMLU and BioASQ does not imply a lack of improvement. 
For example, as shown in Table~\ref{tab:main_results}, the hybrid retriever (used by CoMeta) over MedCPT alone (63.41 $\rightarrow$ 80.49 under Standard Narrative), yet this is not reflected in end-task accuracy due to the ceiling effect, where unoptimized retrieval already provides sufficient context for correct generation. In contrast, on PubMedQA, which requires precise study-level matching, Hit@10 improvement (MedCPT $\rightarrow$ Hybrid: 10.00 $\rightarrow$ 46.67 under Standard Narrative) is accompanied by a +8.33\% end-task accuracy gain, suggesting that retrieval quality 
is a key limiting factor under this setting. Together, these results suggest that the impact of retrieval is conditional on whether it constitutes the bottleneck. \noindent\textit{(2) Blindness to retrieval reliability.} Accuracy metrics cannot reflect retrieval robustness. As shown in Table~\ref{tab:zerohit}, E-Utilities fails on most narrative queries (PubMedQA: 55/60; BioASQ: 76/82), while Adapt-E reduces this to zero. For non-expert users, this constitutes a qualitative reliability shift that is invisible to accuracy-based evaluation. \noindent\textit{(3) Blindness to evidence staleness.} A system may achieve high accuracy by retrieving outdated studies that remain correct for benchmark questions but are unsafe in real-world settings. CoMeta addresses this via metadata-aware filtering in CHSDR, but this safety dimension is orthogonal to accuracy and systematically overlooked by existing benchmarks.

\begin{table}[H]
\centering
\small
\begin{tabular}{lcccc}
\toprule
\textbf{Dataset-Setting} & \textbf{E-utils} & \textbf{Adapt-E (Ours)} \\
\midrule
PubMedQA – Standard  & 22/60 & 0/60 \\
PubMedQA – Narrative & 55/60 & 0/60 \\
BioASQ – Standard    & 18/82 & 0/82 \\
BioASQ – Narrative   & 76/82 & 0/82 \\
\bottomrule
\end{tabular}
\caption{Zero-Hit failure Proportion.}
\label{tab:zerohit}
\end{table}

\noindent\textbf{Adaptive Retrieval Necessity}
Table~\ref{tab:final} reveals that Naive RAG underperforms LLM+CoT on MedQA (67.44 vs.\ 68.60), despite having access to retrieved evidence.  This is consistent with prior findings that RAG can negatively impact the original outcome \cite{wang2023self, asai2023self}. Additionally, Table ~\ref{tab:main_results} also shows the optimal 
retrieval method should be query-specific. While the Hybrid retriever dominates in most settings because it captures complementary evidence missed by symbolic search, it is surpassed by Adapt-E on the PubMedQA Standard Narrative (48.33 vs. 46.67 Hit@10). This exception suggests that for queries derived directly from abstracts, precise lexical matching can be more effective than semantic blending, which might introduce 'semantic noise.' Together, these observations highlight the necessity of adaptive retrieval mechanisms, where the system dynamically determines whether and how to retrieve, rather than a one-size-fits-all solution.

\noindent\textbf{Generalizability of CoMeta} we evaluate Llama-3-8 on both CMMQA and MIRAGE-subset, with Claude-3-Haiku and GPT-4 on CMMQA (Table~\ref{tab:multi_backbone}). We observed two trends: First, CoMeta consistently outperforms Naive RAG across all backbones, with larger gains for smaller models, consistent with the intuition that weaker parametric knowledge increases reliance on retrieval quality. This suggests CoMeta is particularly valuable for smaller open-source models (ideal choices in healthcare due to privacy and cost constraints) while still providing meaningful gains for stronger proprietary models. Second, For Llama-3-8B, the gain (+4.05) on MIRAGE-subset confirms CoMeta’s generalizability beyond oncology. 
\begin{table}[H]
\centering
\small
\setlength{\tabcolsep}{4pt}
\resizebox{\columnwidth}{!}{
\begin{tabular}{ll cc}
\toprule
\textbf{LLM} & \textbf{Method} & \textbf{CMMQA} & \textbf{MIRAGE-subset} \\
\midrule
\multirow{2}{*}{Llama-3-8B}   & Naive RAG & 65.70 & 62.57 \\
                               & \textbf{Ours} & \textbf{70.80} \textcolor{blue}{(+5.10)} & \textbf{66.62} \textcolor{blue}{(+4.05)} \\
\midrule
\multirow{2}{*}{Claude-3-Haiku}& Naive RAG & 69.48 & -- \\
                               & \textbf{Ours} & \textbf{72.39} \textcolor{blue}{(+2.91)} & -- \\
\midrule
\multirow{2}{*}{GPT-4}        & Naive RAG & 75.29 & -- \\
                               & \textbf{Ours} & \textbf{77.90} \textcolor{blue}{(+2.61)} & -- \\
\bottomrule
\end{tabular}
}
\caption{Performance across multiple LLM backbones. MIRAGE-subset:  a stratified 10\% subset of MIRAGE (764 questions, $\geq$50 per sub-dataset). ``--'': omitted due to API budget constraints. We note that MIRAGE-subset results are indicative rather than conclusive given the smaller evaluation scale}
\label{tab:multi_backbone}
\end{table}


\noindent\textbf{Latency and Cost Analysis.}
 E-Utilities latency is 1–2s; semantic search is milliseconds. The main overhead is LLM rewriting. In a "Patient Consultant" scenario, 10–20s latency is acceptable for a reliable, evidence-backed answer. Regarding infrastructure cost (See the system design comparison table in Appendix), CoMeta requires only a static embedding file ($\sim$40GB) plus API access, versus a full BM25+MedCPT index ($\sim$400GB). Furthermore, LLM generation cost also decrease because metadata filtering and SEOS segmentation reduce the reader's context window.

\section{Conclusion}
In this study, we proposed CoMeta, a framework designed for Cancer Patient QA (CPQA). Our work systematically resolves three critical bottlenecks in medical RAG. First, CHSDR overcomes the staleness-semantic dilemma by synergizing real-time, metadata-aware Boolean search (via Adpat-E) with domain-specific semantic retrieval (via MedCPT). Powered by Adapt-E, CHSDR eliminates the ``Zero-Hit" barrier of term-based search, demonstrating exceptional robustness when processing noisy, informal patient narratives. Second, through metadata utilization, CoMeta resolves the retrieval-depth paradox, empirically analyzing the comparative retrieval value of different NCBI sources. Third, SEOS tackles contextual fragmentation by preserving natural clinical dependencies, offering a domain-agnostic chunking strategy that outperforms traditional fixed-length methods. Beyond the observed retrieval performance and end-task accuracy gains (Table \ref{tab:main_results} and Table \ref{tab:final}), our results highlight CoMeta’s particular value in empowering the smaller open-source model (like LLama-3-8B in Table \ref{tab:multi_backbone}). Future work will include exploring advanced semantic segmentation and integrating adaptive retrieval mechanisms as mentioned before. Furthermore, recognizing that oncology care is inherently longitudinal, future work will also investigate evolving CoMeta into a memory-augmented agent system \cite{jiang2026anatomy} deployed on commodity mobile devices \cite{wan2026mobile}, enabling privacy-preserving, long-term health tracking.

\section*{Limitations}

While our evaluation provides robust insights, there are three limitations need to be addressed. First, the reliance on multiple-choice-style evaluation may introduce guessing bias, where random guesses can artificially boost accuracy. Second, the Clinical Narrative queries in CMMQA are synthetic. Although designed to rigorously stress-test the system against worst-case conditions (see Narrative queries Construction Prompt in Appendix A.6), potential distribution shifts relative to real patient queries cannot be entirely ruled out. Finally, existing benchmarks do not explicitly test temporal validity and metadata awareness. Consequently, the real-world safety advantages of our pipeline are underrepresented. We view our current scores as a conservative lower bound on CHSDR’s actual impact in fast-evolving domains like oncology. 


\section*{Acknowledgments}
We would like to thank the anonymous reviewers and the Area Chairs for their constructive feedback, which has significantly improved the quality of this manuscript. The funding of Keeta AI for supporting the travel of this work. 

\bibliography{acl_latex}

@article{nori2023can,
  title={Can Generalist Foundation Models Outcompete Special-Purpose Tuning? Case Study in Medicine},
  author={Nori, Harsha and Lee, Yin Tat and Zhang, Sheng and Carignan, Dean and Edgar, Richard and Fusi, Nicolo and King, Nicholas and Larson, Jonathan and Li, Yuanzhi and Liu, Weishung and others},
  journal={Medicine},
  volume={84},
  number={88.3},
  pages={77--3},
  year={2023}
}

@article{jeong2024improving,
  title={Improving medical reasoning through retrieval and self-reflection with retrieval-augmented large language models},
  author={Jeong, Minbyul and Sohn, Jiwoong and Sung, Mujeen and Kang, Jaewoo},
  journal={Bioinformatics},
  volume={40},
  number={Supplement\_1},
  pages={i119--i129},
  year={2024},
  publisher={Oxford University Press}
}

@article{singhal2023large,
  title={Large language models encode clinical knowledge},
  author={Singhal, Karan and Azizi, Shekoofeh and Tu, Tao and Mahdavi, S Sara and Wei, Jason and Chung, Hyung Won and Scales, Nathan and Tanwani, Ajay and Cole-Lewis, Heather and Pfohl, Stephen and others},
  journal={Nature},
  volume={620},
  number={7972},
  pages={172--180},
  year={2023},
  publisher={Nature Publishing Group UK London}
}

@article{lievin2024can,
  title={Can large language models reason about medical questions?},
  author={Li{\'e}vin, Valentin and Hother, Christoffer Egeberg and Motzfeldt, Andreas Geert and Winther, Ole},
  journal={Patterns},
  volume={5},
  number={3},
  year={2024},
  publisher={Elsevier}
}

@misc{xu2026selfcorrectingrag,
      title={Self-Correcting RAG: Enhancing Faithfulness via MMKP Context Selection and NLI-Guided MCTS}, 
      author={Shijia Xu and Zhou Wu and Xiaolong Jia and Yu Wang and Kai Liu and April Xiaowen Dong},
      year={2026},
      eprint={2604.10734},
      archivePrefix={arXiv},
      primaryClass={cs.CL},
      url={https://arxiv.org/abs/2604.10734}, 
}

@article{wan2026mobile,
  title={Mobile-VTON: High-Fidelity On-Device Virtual Try-On},
  author={Wan, Zhenchen and Chen, Ce and Lin, Runqi and Huang, Jiaxin and Chen, Tianxi and Xu, Yanwu and Liu, Tongliang and Gong, Mingming},
  journal={arXiv e-prints},
  pages={arXiv--2603},
  year={2026}
}

@article{jiang2026anatomy,
  title={Anatomy of Agentic Memory: Taxonomy and Empirical Analysis of Evaluation and System Limitations},
  author={Jiang, Dongming and Li, Yi and Wei, Songtao and Yang, Jinxin and Kishore, Ayushi and Zhao, Alysa and Kang, Dingyi and Hu, Xu and Chen, Feng and Li, Qiannan and others},
  journal={arXiv preprint arXiv:2602.19320},
  year={2026}
}

@inproceedings{xiong-etal-2024-benchmarking,
    title = "Benchmarking Retrieval-Augmented Generation for Medicine",
    author = "Xiong, Guangzhi  and
      Jin, Qiao  and
      Lu, Zhiyong  and
      Zhang, Aidong",
    editor = "Ku, Lun-Wei  and
      Martins, Andre  and
      Srikumar, Vivek",
    booktitle = "Findings of the Association for Computational Linguistics: ACL 2024",
    month = aug,
    year = "2024",
    address = "Bangkok, Thailand",
    publisher = "Association for Computational Linguistics",
    url = "https://aclanthology.org/2024.findings-acl.372",
    doi = "10.18653/v1/2024.findings-acl.372",
    pages = "6233--6251",
}

@article{liang2025rgar,
  title={RGAR: Recurrence Generation-augmented Retrieval for Factual-aware Medical Question Answering},
  author={Liang, Sichu and Zhang, Linhai and Zhu, Hongyu and Wang, Wenwen and He, Yulan and Zhou, Deyu},
  journal={arXiv preprint arXiv:2502.13361},
  year={2025}
}

@article{tsatsaronis2015overview,
  title={An overview of the BIOASQ large-scale biomedical semantic indexing and question answering competition},
  author={Tsatsaronis, George and Balikas, Georgios and Malakasiotis, Prodromos and Partalas, Ioannis and Zschunke, Matthias and Alvers, Michael R and Weissenborn, Dirk and Krithara, Anastasia and Petridis, Sergios and Polychronopoulos, Dimitris and others},
  journal={BMC bioinformatics},
  volume={16},
  number={1},
  pages={138},
  year={2015},
  publisher={Springer}
}

@article{umapathi2023med,
  title={Med-halt: Medical domain hallucination test for large language models},
  author={Umapathi, Logesh Kumar and Pal, Ankit and Sankarasubbu, Malaikannan},
  journal={arXiv preprint arXiv:2307.15343},
  year={2023}
}

@article{ji2023survey,
  title={Survey of hallucination in natural language generation},
  author={Ji, Ziwei and Lee, Nayeon and Frieske, Rita and Yu, Tiezheng and Su, Dan and Xu, Yan and Ishii, Etsuko and Bang, Ye Jin and Madotto, Andrea and Fung, Pascale},
  journal={ACM Computing Surveys},
  volume={55},
  number={12},
  pages={1--38},
  year={2023},
  publisher={ACM New York, NY}
}

@article{lewis2020retrieval,
  title={Retrieval-augmented generation for knowledge-intensive nlp tasks},
  author={Lewis, Patrick and Perez, Ethan and Piktus, Aleksandra and Petroni, Fabio and Karpukhin, Vladimir and Goyal, Naman and K{\"u}ttler, Heinrich and Lewis, Mike and Yih, Wen-tau and Rockt{\"a}schel, Tim and others},
  journal={Advances in Neural Information Processing Systems},
  volume={33},
  pages={9459--9474},
  year={2020}
}

@article{gao2023retrieval,
  title={Retrieval-augmented generation for large language models: A survey},
}

@article{singhal2023towards,
  title={Towards expert-level medical question answering with large language models},
  author={Singhal, Karan and Tu, Tao and Gottweis, Juraj and Sayres, Rory and Wulczyn, Ellery and Hou, Le and Clark, Kevin and Pfohl, Stephen and Cole-Lewis, Heather and Neal, Darlene and others},
  journal={arXiv preprint arXiv:2305.09617},
  year={2023}
}

@article{liu2022llamaindex,
  title={LlamaIndex},
  author={Liu, Jerry},
  journal={Acceso el},
  volume={6},
  year={2022}
}

@article{wei2022chain,
  title={Chain-of-thought prompting elicits reasoning in large language models},
  author={Wei, Jason and Wang, Xuezhi and Schuurmans, Dale and Bosma, Maarten and Xia, Fei and Chi, Ed and Le, Quoc V and Zhou, Denny and others},
  journal={Advances in neural information processing systems},
  volume={35},
  pages={24824--24837},
  year={2022}
}

@article{robertson2009probabilistic,
  title={The probabilistic relevance framework: BM25 and beyond},
  author={Robertson, Stephen and Zaragoza, Hugo and others},
  journal={Foundations and Trends{\textregistered} in Information Retrieval},
  volume={3},
  number={4},
  pages={333--389},
  year={2009},
  publisher={Now Publishers, Inc.}
}

@article{muennighoff2022mteb,
    doi = {10.48550/ARXIV.2210.07316},
    url = {https://arxiv.org/abs/2210.07316},
    author = {Muennighoff, Niklas and Tazi, Nouamane and Magne, Lo{\"\i}c and Reimers, Nils},
    title = {MTEB: Massive Text Embedding Benchmark},
    publisher = {arXiv},
    journal={arXiv preprint arXiv:2210.07316},  
    year = {2022}
}

@article{miao2025improving,
  title={Improving Large Language Model Applications in the Medical and Nursing Domains With Retrieval-Augmented Generation: Scoping Review},
  author={Miao, Yiqun and Zhao, Yuhan and Luo, Yuan and Wang, Huiying and Wu, Ying},
  journal={Journal of Medical Internet Research},
  volume={27},
  pages={e80557},
  year={2025},
  publisher={JMIR Publications Toronto, Canada}
}

@article{jin2023medcpt,
  title={MedCPT: Contrastive Pre-trained Transformers with large-scale PubMed search logs for zero-shot biomedical information retrieval},
  author={Jin, Qiao and Kim, Won and Chen, Qingyu and Comeau, Donald C and Yeganova, Lana and Wilbur, W John and Lu, Zhiyong},
  journal={Bioinformatics},
  volume={39},
  number={11},
  pages={btad651},
  year={2023},
  publisher={Oxford University Press}
}

@article{wang2019multi,
  title={Multi-passage bert: A globally normalized bert model for open-domain question answering},
  author={Wang, Zhiguo and Ng, Patrick and Ma, Xiaofei and Nallapati, Ramesh and Xiang, Bing},
  journal={arXiv preprint arXiv:1908.08167},
  year={2019}
}

@article{tang2024multihop,
  title={MultiHop-RAG: Benchmarking Retrieval-Augmented Generation for Multi-Hop Queries},
  author={Tang, Yixuan and Yang, Yi},
  journal={arXiv e-prints},
  pages={arXiv--2401},
  year={2024}
}

@article{hearst1997text,
  title={Text tiling: Segmenting text into multi-paragraph subtopic passages},
  author={Hearst, Marti A},
  journal={Computational linguistics},
  volume={23},
  number={1},
  pages={33--64},
  year={1997}
}

@inproceedings{hendrycks2020measuring,
  title={Measuring Massive Multitask Language Understanding},
  author={Hendrycks, Dan and Burns, Collin and Basart, Steven and Zou, Andy and Mazeika, Mantas and Song, Dawn and Steinhardt, Jacob},
  booktitle={International Conference on Learning Representations},
  year={2020}
}

@article{jin2021disease,
  title={What Disease Does This Patient Have? A Large-Scale Open Domain Question Answering Dataset from Medical Exams},
  author={Jin, Di and Pan, Eileen and Oufattole, Nassim and Wei-Hung, Weng and Fang, Hanyi and Szolovits, Peter},
  journal={Applied Sciences},
  volume={11},
  number={14},
  pages={6421},
  year={2021},
  publisher={MDPI AG}
}

@inproceedings{pal2022medmcqa,
  title={Medmcqa: A large-scale multi-subject multi-choice dataset for medical domain question answering},
  author={Pal, Ankit and Umapathi, Logesh Kumar and Sankarasubbu, Malaikannan},
  booktitle={Conference on health, inference, and learning},
  pages={248--260},
  year={2022},
  organization={PMLR}
}

@inproceedings{jin2019pubmedqa,
  title={PubMedQA: A Dataset for Biomedical Research Question Answering},
  author={Jin, Qiao and Dhingra, Bhuwan and Liu, Zhengping and Cohen, William and Lu, Xinghua},
  booktitle={Proceedings of the 2019 Conference on Empirical Methods in Natural Language Processing and the 9th International Joint Conference on Natural Language Processing (EMNLP-IJCNLP)},
  pages={2567--2577},
  year={2019}
}

@incollection{kans2024entrez,
  title={Entrez direct: E-utilities on the UNIX command line},
  author={Kans, Jonathan},
  booktitle={Entrez programming utilities help [Internet]},
  year={2024},
  publisher={National Center for Biotechnology Information (US)}
}

@misc{li2023making,
      title={Making Large Language Models A Better Foundation For Dense Retrieval}, 
      author={Chaofan Li and Zheng Liu and Shitao Xiao and Yingxia Shao},
      year={2023},
      eprint={2312.15503},
      archivePrefix={arXiv},
      primaryClass={cs.CL}
}

@misc{pubmedbert,
  author = {Yu Gu and Robert Tinn and Hao Cheng and Michael Lucas and Naoto Usuyama and Xiaodong Liu and Tristan Naumann and Jianfeng Gao and Hoifung Poon},
  title = {Domain-Specific Language Model Pretraining for Biomedical Natural Language Processing},
  year = {2020},
  eprint = {arXiv:2007.15779},
}

@inproceedings{cormack2009reciprocal,
  title={Reciprocal rank fusion outperforms condorcet and individual rank learning methods},
  author={Cormack, Gordon V and Clarke, Charles LA and Buettcher, Stefan},
  booktitle={Proceedings of the 32nd international ACM SIGIR conference on Research and development in information retrieval},
  pages={758--759},
  year={2009}
}

@article{truhn2023large,
  title={Large language models should be used as scientific reasoning engines, not knowledge databases},
  author={Truhn, Daniel and Reis-Filho, Jorge S and Kather, Jakob Nikolas},
  journal={Nature medicine},
  volume={29},
  number={12},
  pages={2983--2984},
  year={2023},
  publisher={Nature Publishing Group US New York}
}

@inproceedings{asai2023self,
  title={Self-RAG: Learning to Retrieve, Generate, and Critique through Self-Reflection},
  author={Asai, Akari and Wu, Zeqiu and Wang, Yizhong and Sil, Avirup and Hajishirzi, Hannaneh},
  booktitle={The Twelfth International Conference on Learning Representations},
  year={2023}
}

@article{landhuis2016scientific,
  title={Scientific literature: Information overload},
  author={Landhuis, Esther},
  journal={Nature},
  volume={535},
  number={7612},
  pages={457--458},
  year={2016},
  publisher={Nature Publishing Group UK London}
}

@article{lu2019spell,
  title={Spell checker for consumer language (CSpell)},
  author={Lu, Chris J and Aronson, Alan R and Shooshan, Sonya E and Demner-Fushman, Dina},
  journal={Journal of the American Medical Informatics Association},
  volume={26},
  number={3},
  pages={211--218},
  year={2019},
  publisher={Oxford University Press}
}

@incollection{abacha2019bridging,
  title={Bridging the gap between consumers’ medication questions and trusted answers},
  author={Abacha, Asma Ben and Mrabet, Yassine and Sharp, Mark and Goodwin, Travis R and Shooshan, Sonya E and Demner-Fushman, Dina},
  booktitle={MEDINFO 2019: Health and Wellbeing e-Networks for All},
  pages={25--29},
  year={2019},
  publisher={IOS Press}
}

@inproceedings{jiang2023boot,
  title={Boot and Switch: Alternating Distillation for Zero-Shot Dense Retrieval},
  author={Jiang, Fan and Xu, Qiongkai and Drummond, Tom and Cohn, Trevor},
  booktitle={The 2023 Conference on Empirical Methods in Natural Language Processing}
}

@inproceedings{wang2023self,
  title={Self-Knowledge Guided Retrieval Augmentation for Large Language Models},
  author={Wang, Yile and Li, Peng and Sun, Maosong and Liu, Yang},
  booktitle={Findings of the Association for Computational Linguistics: EMNLP 2023},
  pages={10303--10315},
  year={2023}
}

@inproceedings{wang2025trustworthy,
  title={Trustworthy Medical Question Answering: An Evaluation-Centric Survey},
  author={Wang, Yinuo and Wang, Baiyang and Mercer, Robert and Rudzicz, Frank and Roy, Sudipta Singha and Ren, Pengjie and Chen, Zhumin and Wang, Xindi},
  booktitle={Proceedings of the 2025 Conference on Empirical Methods in Natural Language Processing},
  pages={27477--27490},
  year={2025}
}

\appendix
\clearpage
\section{Example Appendix}
\label{sec:appendix}
\subsection{Test Data Details} 
We construct evaluation dataset by appling a MeSH-based filter on six widely used medical datasets:
\begin{itemize}
    \item PubMedQA \cite{jin2019pubmedqa} and BioASQ \cite{xiong-etal-2024-benchmarking} are QA datasets with answers from biomedical literature. PubMedQA  (500 questions) is characterized by \textbf{single-document grounding}, where each manually annotated question is paired with a single corresponding PMID. In contrast, \textbf{BioASQ} \cite{xiong-etal-2024-benchmarking} (618 questions) features \textbf{multi-document grounding}, providing a list of multiple relevant PMIDs for each question. They are suitable for  document retrieval evaluation due to varying evidence densities and the explicit annotation of gold-truth PMIDs. 
    \item MedQA \cite{jin2021disease}, MedMCQA \cite{pal2022medmcqa}, and the medical subsets in MMLU \cite{hendrycks2020measuring} collected questions from medical exams. Specifically, MedQA \cite{jin2021disease} contains 1273 questions from the US Medical Licensing Examination (USMLE), focusing on complex clinical decision-making scenarios faced by professionals. MedMCQA is a large-scale dataset covering 21 medical subjects with 4183-question development set, focusing on evaluating across diverse healthcare topics. MMLU-med is a subset of the MMLU benchmark comprising 1,089 questions across six subtasks (e.g., professional medicine, human genetics), serving as a testbed for general biomedical reasoning. 

    \item HealthSearchQA \cite{singhal2023towards}, comprising general consumer search queries, lacks definitive answers and is therefore unsuitable for accuracy assessment. However, its question lengths make it suitable for investigating the impact of question length on document retrieval methods.
\cite{nori2023can,singhal2023large,lievin2024can}.
\end{itemize}
First five datasets are multiple-choice QA datasets, because using multiple-choice questions can simplify evaluation, eliminate biases from text similarity computations or human annotation, and align with large-scale medical QA systems evaluations. 

\subsection{The system design comparison Table}
\label{sec:bm25_vs_eutils}
BM25 and E-Utilities address fundamentally different problems within the RAG pipeline; they are not simply interchangeable sparse retrievers. Actually, our target deployment is real-world clinical settings that require metadata-aware, up-to-date evidence access without maintaining massive local PubMed mirrors. Thus, a static BM25+MedCPT index is not a suitable baseline. Table \ref{tab:bm25_vs_eutils} shows the key distinctions in indexing paradigms. 

\begin{table}[h]
\centering
\small
\renewcommand{\arraystretch}{1.3}
\begin{tabular}{@{} >{\raggedright\arraybackslash}p{0.4\linewidth} 
                    >{\raggedright\arraybackslash}p{0.26\linewidth} 
                    >{\raggedright\arraybackslash}p{0.28\linewidth} @{}}
\toprule
\textbf{Feature} & \textbf{BM25 + MedCPT} & \textbf{E-Utils + MedCPT} \\
\midrule
\textbf{Lexical Complementarity} & $\checkmark$ (Strong) & $\triangle$ (Moderate) \\
\textbf{Query Expansion} \newline (ATM $\rightarrow$ MeSH) & $\times$ & $\checkmark$ \\
\textbf{Metadata-Awareness} & $\times$ & $\checkmark$ \\
\textbf{Real-time Access} & $\times$ & $\checkmark$ \\
\textbf{Infrastructure} & $\sim$400GB index & $\sim$40GB index \newline + Live API \\
\bottomrule
\end{tabular}
\caption{System design comparison.}
\label{tab:bm25_vs_eutils}
\end{table}

\subsection{The representative failure-recovery illustration}

\label{sec:case_study}

\begin{table}[H]
\centering
\small
\renewcommand{\arraystretch}{1.4}
\begin{tabular}{p{0.18\linewidth} p{0.68\linewidth} c}
\toprule
\textbf{Phase} & \textbf{Content / Query Candidates} & \textbf{Hits} \\
\midrule
\textbf{Patient Query} & \textit{``...The doc thinks she might have a tumor in her pancreas and they're gonna do a biopsy tomorrow. My sister and I are her caregivers and we think it's best if we don't tell her if it's canc'r...''} & \textbf{0} \\
\midrule
\textbf{Candidate 1} (Strict) & \texttt{"pancreatic tumor" AND "biopsy" AND "truth-telling" AND "caregivers"} & 0 \\
\textbf{Candidate 2} & \texttt{"pancreatic tumor" AND "biopsy" AND ("truth-telling" OR "disclosure")} & 10 \\
\textbf{Candidate 3} & \texttt{"pancreatic tumor" AND "biopsy" AND "caregivers"} & 10 \\
\textbf{Candidate 4} & \texttt{"pancreatic tumor" AND "caregivers" AND ("truth-telling" OR "bad news")} & 10 \\
\textbf{Candidate 5} (Relaxed)& \texttt{"pancreatic tumor" OR "biopsy" OR "truth-telling"} & 32 \\
\bottomrule
\end{tabular}
\caption{A representative failure-recovery case demonstrating the Adaptive Fallback mechanism. The system sequentially relaxes the Boolean logic until sufficient evidence is retrieved, efficiently overcoming the "Zero-Hit" barrier typical of raw sparse retrieval.}
\label{tab:fallback_case}
\end{table}

To illustrate how the Adapt-E module achieves a near-zero Hit0 rate on complex patient narratives, Table \ref{tab:fallback_case} presents a representative failure-recovery scenario. When a standard query fails, the system does not fail outright. Instead, it utilizes the LLM to generate a hierarchy of Boolean search strings, progressing from strict clinical constraints to relaxed conceptual matching. The system executes these candidates sequentially and triggers an early-stopping mechanism once sufficient documents are retrieved (e.g., stopping at Candidate 2 or 3). This mechanism avoids unnecessary relaxation and preserves retrieval precision while ensuring high robustness against noisy inputs.

\subsection{The pseudo-code of SEOS}
\label{pseudo}

Algorithm \ref{alg:seos} outlines the process of Semantic Enhanced Overlap Segmentation (SEOS). This method  ensures that the document is split at natural semantic inflection points (where major topic shifts occur), while our sentence-level overlapping strategy guarantees that logical continuity is preserved across chunks.

\begin{algorithm}[H]
\caption{{S}emantic \textbf{E}nhanced \textbf{O}verlap \textbf{S}egmentation}
\label{alg:seos} 
\begin{algorithmic}[1]
\Procedure{SEOS}{document, k, \textit{embed\_model}}
    \State Split document into sentences using sentence\_splitter
    \State Merge each sentence with its \(k\) neighboring sentences
    \State Get each sentence group's embedding by \textit{embed\_model}
    \State Use cosine dissimilarity change between adjacent groups to identify similarity inflection (semantic shift) points
    \State Adjust chunk\_size based on \textit{embed\_model}
    \State \(n \gets \text{len}(document) / chunk\_size\)
    \State Breakpoints: \text{top \(n\) points with largest semantic shift}
    \State Split document into chunks at breakpoints
    \State Overlap is added to per chunk by including the first sentence of the next chunk to preserve sequence order
    \State \Return Chunks
\EndProcedure
\end{algorithmic}
\end{algorithm}

\subsection{Retrieval Examples}

\noindent
\begin{PromptBox}{Retrieval Examples (Output)}
\begin{Verbatim}[fontsize=\small, breaklines, breakanywhere]
"step_by_step_thinking": 
"The question asks whether pneumologists' skepticism about chemotherapy and survival in advanced non-smallcell lung carcinoma is justified. The context information suggests that chemotherapy can prolong survival and improve quality of life for patients with advanced non-small cell lung cancer, especially those with good performance status. Several studies have shown that platinum-based chemotherapy regimens are the standard of care and can lead to improved survival compared to supportive care alone. 

However, the context also suggests that there may be some skepticism among pneumologists about the benefits of chemotherapy.
The context indicates that while chemotherapy can prolong survival, the benefits are moderate, with median and 1-year survival of 8--10 months and 30--40\%, respectively. Additionally, the survival benefits vary depending on factors like disease stage, patient characteristics, and chemotherapy regimen. 

Overall, the context information suggests that chemotherapy can provide meaningful survival and quality of life benefits for many patients with advanced non-small cell lung cancer, but the benefits are not dramatic and must be weighed against the potential for toxicity. This may explain some of the skepticism among pneumologists.

Based on the information provided, I believe the answer is B: no, the pneumologists' skepticism is not justified. While the benefits of chemotherapy are not as large as some might hope, the context information indicates that it can still provide important survival and quality of life benefits for many patients, and the risks are generally manageable. The skepticism may be understandable, but the evidence does not seem to fully support it."

"answer_choice": "B"
\end{Verbatim}
\end{PromptBox}

\subsection{LLM Prompts} 
\clearpage





\begin{figure*}[p]
\begin{PromptBox}{Prompt: Rewrite to Clinical Narrative Queries}
\begin{Verbatim}[fontsize=\small]
prompt_searchapi = """
### Role
You are a simulator for a Consumer Cancer QA System. Real users have no medical background, including cancer patients and their family members. 

### Input Data
RAW QUESTION: {question}

### Task
Identify the medical scenario, You need to transform RAW QUESTION into a a first-person query from a real user.

### Requirements
1. Use layperson language (e.g., "spread" instead of "metastasize").
2. Include 1-2 minor typos (e.g., "cancr" for "cancer") or grammar slips.
3. Considering the real situation of non-expert users, rewrite from patient perspective.
4. Keep the core medical doubt unchanged so it can still be answered.

### Output Format
Return ONLY the rewritten string. No explanations.
 
"""
\end{Verbatim}
\end{PromptBox}
\label{CMMQA-rewrite}
\end{figure*}

\begin{figure*}[p]
\begin{PromptBox}{LLM with CoT + RAG}
\begin{Verbatim}[fontsize=\small]
prompt_template = """
You are a helpful medical expert. 

QUESTION: {question}
POTENTIAL CHOICES: {options}
EVIDENCE: {context} 

Task: answer a multi-choice medical question with provided evidence. 

Instructions:
1.think step-by-step then choose the answer for QUESTION from the POTENTIAL CHOICES, using the EVIDENCE. 
2.Organize your output in a json formatted as Dict{{"step_by_step_thinking":Str(explanation), "answer_choice": Str{{A/B/C/...}}}}.

Note: 
1.Only Dict in output is needed. No need for other text.
2.Your responses will be used for research purposes only, so please have a definite answer.
3.Keep your answer ground in the facts of the EVIDENCE.
"""
\end{Verbatim}
\end{PromptBox}
\end{figure*}

\begin{figure*}[p]
\begin{PromptBox}{Prompt: Rewrite for boolean queries and set time constraints}
\begin{Verbatim}[fontsize=\small]
prompt_searchapi = """
You are a helpful expert in information extraction.

QUERY: {query}

Task: generate boolean search queries for QUERY

Instructions: 
1. Preprocessing Stage: 
- **Correct** any orthographic or grammatical errors in the QUERY.
- **Analyze** the Question Type and User intent to identify core concepts and **SEPARATE** the "Noise" from the "Core Dilemma".
- **Extract** the core terms. Especially identify **PICO** (Population, Intervention, Outcome)
- **Abstraction** specific scenario details into general medical concepts (e,g. "Don't tell" -> "Truth-telling").
- **Time**: If time constraints (like "recent") appears, set reasonable year_range (e.g., 5). Otherwise None.
2. Boolean Queries Generation Stage: 
- Provide 5 possible boolean search queries using core terms.
- Start with the most specific query (Strict AND of all terms). 
- Progressively relax the strict query based on your Intent Analysis in stage 1:          
[Strategy A (Synonyms): If a term is a CORE concept, relax it by adding synonyms];
[Strategy B (Pruning): If a term is a SPECIFIC but NOt CORE detail (e.g., "in Italy"), remove] 
[Strategy C (Logical Relaxation)]: Switch AND to OR between distinct core concepts


Examples:
# Example 1: 
Input: "prevalence of concurrent diabetes and cardiovascular disease in women in Beijing"
Output: (
  [
    '"prevalence" AND "diabetes" AND "cardiovascular disease" AND "women" AND "Beijing"',
    '"prevalence" AND "diabetes" AND "heart disease" AND "women"', 
    '"epidemiology" AND "concurrent" AND "diabetes" AND "cardiovascular disease"',
    '"diabetes" AND "cardiovascular disease" AND "comorbidity"',
    '"prevalence" AND ("diabetes" OR "cardiovascular disease")'
  ],
None
)
# Example 2: 
# Abstractions: Stroke->End of life; Disconnect->Withdrawal; Son/Daughter->Family Conflict
Input: "75-year-old man with severe stroke and ventilator. Daughter wants to disconnect the machine but son disagrees."
Output: (
  [
      '"withdrawal of life support" AND "family conflict" AND "surrogate decision making"', 
      '"withdrawal of life support" AND ("family conflict" OR "disagreement") AND "decision making"',   
      '"withdrawal of life support" AND "family conflict"',                                            
      '"surrogate decision making" AND "family conflict"',                                            
      '"withdrawal of life support" OR "end of life care"'   
  ],
None
)
# Example 3: Time Sensitive
Input: "latest treatments for glioblastoma in the last 8 years"
Output: (
    [
        '"Glioblastoma"[MeSH] AND "Therapeutics"[MeSH]',
        '"Glioblastoma" AND "Treatment" AND "Novel"',
        '"Brain Neoplasms" AND "Therapy"',
        '"Glioblastoma" AND "Immunotherapy"',
        '"Glioblastoma" OR "Brain Tumor"'
    ],
    8
)

Note: Return ONLY the Python tuple (list_of_strings, year_range_int). No other text.

"""
\end{Verbatim}
\end{PromptBox}
\end{figure*}

\end{document}